\pdfoutput=1
\documentclass[11pt]{article}

\usepackage[preprint]{acl}

\usepackage{times}
\usepackage{latexsym}

\usepackage[T1]{fontenc}
\usepackage[utf8]{inputenc}

\usepackage{microtype}

\usepackage{inconsolata}

\usepackage{graphicx}

\usepackage[shortlabels]{enumitem}

\usepackage{array}
\usepackage{booktabs}
\usepackage{multirow}
\usepackage[normalem]{ulem}
\usepackage{multirow}  % Required for multirow functionality
\usepackage{listings}
\usepackage{xcolor}

\newif\ifcorrections % Define a conditional
\correctionstrue% Set the mode to final by default. Use \correctionsfalse for final mode.

\definecolor{codegreen}{rgb}{0,0.6,0}
\definecolor{codegray}{rgb}{0.5,0.5,0.5}
\definecolor{codepurple}{rgb}{0.58,0,0.82}
\definecolor{backcolour}{rgb}{0.94,0.94,0.94}

\lstdefinestyle{mystyle}{   
    backgroundcolor=\color{backcolour},  
    framexleftmargin=1mm,
    framexrightmargin=1mm,
    xleftmargin=1mm,
    xrightmargin=1mm,
    basicstyle=\ttfamily\scriptsize,
    breakatwhitespace=true,         
    breaklines=true,                 
    captionpos=b,        
    keepspaces=true,                
    showspaces=false,                
    showstringspaces=false,
    showtabs=false,                  
}

\title{Can LLMs Predict Citation Intent? An Experimental Analysis of In-context Learning and Fine-tuning on Open LLMs}

\author{Paris Koloveas \\
  IMSI, ATHENA RC / Athens, GR \\
  University of the Peloponnese / Tripolis, GR \\
  \texttt{pkoloveas@athenarc.gr} \\\And
  Serafeim Chatzopoulos \\
  IMSI, ATHENA RC / Athens, GR \\
  \texttt{schatz@athenarc.gr} \\\AND
  Thanasis Vergoulis \\
  IMSI, ATHENA RC / Athens, GR \\
  \texttt{vergoulis@athenarc.gr} \\\And
  Christos Tryfonopoulos \\
  University of the Peloponnese / Tripolis, GR \\
  \texttt{trifon@uop.gr} \\}

\begin{document}
\maketitle
\begin{abstract} 
    This work investigates the ability of open Large Language Models (LLMs) to predict citation intent through in-context learning and fine-tuning. Unlike traditional approaches relying on domain-specific pre-trained models like SciBERT, we demonstrate that general-purpose LLMs can be adapted to this task with minimal task-specific data. We evaluate twelve model variations across five prominent open LLM families using zero-, one-, few-, and many-shot prompting. Our experimental study identifies the top-performing model and prompting parameters through extensive in-context learning experiments. We then demonstrate the significant impact of task-specific adaptation by fine-tuning this model, achieving a relative F1-score improvement of 8\% on the SciCite dataset and 4.3\% on the ACL-ARC dataset compared to the instruction-tuned baseline. These findings provide valuable insights for model selection and prompt engineering. Additionally, we make our end-to-end evaluation framework and models openly available for future use.
\end{abstract}

Citations are references of research articles to external sources of information included to support claims, provide context, criticize, or acknowledge prior work. Although their primary function is to inform and redirect the reader, citations are also frequently used for other purposes, such as serving as proxies of scientific impact in various types of analysis~\citep{KanellosVSDV21}. In such cases, understanding the exact intent of a citation is crucial, as not all types of citations should be considered. For instance, while measuring the impact of an article, citations that criticize the work should generally not be considered as contributing to its impact.

Predicting citation intent based on its context (i.e., the sentences in the manuscript that accompany the citation) and other related information, has become an important classification problem to support the aforementioned use cases. Existing approaches have traditionally relied on linguistic features~\citep{JKH2018}, machine learning methods~\citep{TST2006a}, and, in recent years, on domain-specific pre-trained language models (PLMs), such as SciBERT~\citep{BLC2019}, that require large scientific datasets such as~\citep{s2orc} and task-specific architectures\footnote{In this paper, when referring to traditional PLMs in this context, we primarily mean models based on the encoder-only Transformer architecture, such as BERT and its derivatives (e.g., SciBERT), which were commonly fine-tuned for specific classification tasks. This distinguishes them from the general-purpose Large Language Models (LLMs) evaluated in this work, which typically utilize decoder-only or encoder-decoder Transformer architectures and are often leveraged for their generative and in-context learning capabilities.}. In contrast, this work is the first to explore the potential of open, general-purpose Large Language Models (LLMs) 
-- which, despite potentially encountering scientific text during their broad pre-training, are not specifically optimized for the scientific domain like SciBERT -- 
to accurately identify citation intent,
% without using any scientific-trained PLMs, 
evaluating their effectiveness through in-context learning and fine-tuning on minimal task-specific data. 

% Large Language Models (LLMs) are artificial intelligence (AI) approaches trained on vast amounts of text data to understand and generate human-like language. They use deep learning, particularly transformer architectures, to process and generate text, making them highly effective for various tasks such as translation and summarization. Over time, LLMs have evolved into accessible tools for tackling machine learning problems without requiring extensive expertise in AI. Their key strengths include ease of use, as they can be accessed via APIs or chat interfaces and versatility, since they can handle a wide range of tasks without task-specific training. Clearly, exploring their potential in citation intent classification presents an intriguing and valuable research opportunity. 

Large Language Models are advanced natural language processing systems trained on extensive text corpora to perform a wide range of language tasks. Unlike traditional pre-trained language models, LLMs are general-purpose models capable of adapting to new tasks with minimal additional training. Their ability to process and generate coherent text across diverse contexts makes them particularly suitable for tasks like citation intent classification, where understanding nuanced language patterns is essential.

In-context learning is a paradigm that allows LLMs to learn tasks given only a few examples in the form of demonstration~\citep{BMR2020,icl-survey}. In-context learning is particularly suited to citation intent classification, as it allows the models to leverage their contextual understanding of language to make accurate predictions.

In our study, we conduct an extensive experimental analysis of $12$ general-purpose Instruction-tuned LLMs from $5$ model families,  on two widely used datasets for this task. In this context, we analyze the impact of multiple parameters on model performance and identify the optimal configurations and best-performing models for our problem. These experiments address the following research questions: 

\begin{itemize}[nosep, topsep=1pt]
    \item \textit{RQ1:} How well can pre-trained LLMs perform on citation intent classification without task-specific training? 
    \item \textit{RQ2:} What are the differences in performance between open LLMs of varying parameter counts?
    \item \textit{RQ3:} How do different prompting-related parameters affect model performance? 
\end{itemize}

%\begin{table}[b]
%    \centering
%    \footnotesize
%    \begin{tabular}{lp{0.8\linewidth}}
%      \hline
%                     & \textbf{Question}  \\
%      \hline
%      RQ1 & How well can pre-trained LLMs perform on citation intent classification without task-specific training compared to the state-of-the-art? \\
%      \hline
%      RQ2 & What are the differences in performance between small and mid-sized state-of-the-art open LLMs?\\
%      \hline
%      RQ3 & How do different prompting-related parameters affect model performance? Are there definitive optimal choices for some parameters?\\
%      \hline
%      RQ4 & How much does supervised fine-tuning with task-specific training increase the performance of the instruction-tuned models?\\
%      \hline
%      \end{tabular}
%          \caption{Research questions.}\label{tab:rqs}
%      \end{table}

Furthermore, we take the analysis a step further by selecting the top-performing cases from the aforementioned experiments for Supervised Fine-Tuning (SFT) on our datasets, allowing us to evaluate their optimal performance. This essentially addresses an additional research question:

\begin{itemize}
    \item \textit{RQ4:} How much does supervised fine-tuning with task-specific training affect the performance of the instruction-tuned models?
\end{itemize}

Our experiments  aim to contribute to the broader research community by guiding prompt engineering and model selection strategies for similar tasks. To further assist researchers, we openly\footnote{\url{https://github.com/athenarc/CitationIntentOpenLLM}} provide our complete testing suite and evaluation results, allowing seamless integration of Hugging Face models and easy configuration adjustments. We also publish the weights and checkpoints of the fine-tuned models\footnote{\url{https://huggingface.co/collections/sknow-lab/citationintentllm-67b72f1d5ca6113f960dba04}} presented in Section~\ref{sec:fine_tuning}.

\section{Methodology}\label{sec:experimental_setup}
% \section{\ser{Experimental Setup and}{Evaluation} Methodology}\label{sec:experimental_setup}

% \emph{\color{red} \textbf{Paris:} Should write something introductory here - probably merged froom Section 3.1 below.}

% \subsection{Overview and Objectives}\label{sec:overview_objectives}
% \emph{\color{red} \textbf{Paris:} Might merge this and 3.1 with above.}

In this section, we elaborate on the evaluation methodology followed in our experimental study. We first provide an overview of the used models (Section~\ref{sec:models}), datasets (Section~\ref{sec:datasets}), and configuration parameters (Section~\ref{sec:configuration-params}). Finally, in Section~\ref{sec:technical_specifications}, we present the technical specifications of the system we used for the experimentation.

\subsection{Models}\label{sec:models}

For our experiments, we selected the Instruction-tuned versions of five prominent open-weight model families: \textit{LLaMA}~\citep{llama3}, \textit{Mistral}~\citep{mistral}, \textit{Phi}~\citep{phi3}, \textit{Gemma}~\citep{gemma2}, and \textit{Qwen}~\citep{qwen2,qwen2_5}. 
% For several models, we test different variations of training parameters, resulting to a total of $12$ different models.
% The experimental design was shaped by practical constraints and research goals. 
A key challenge was performing the experiments on commodity hardware, both due to computational limitations and to demonstrate that citation intent classification can be executed efficiently with limited resources. This influenced the selection of models, which range in size from small ($1$B parameters) to medium-sized ($27$B parameters).
% \ser{The design also sought to maximize the range of configurations explored to identify optimal performance scenarios.}{}

% \ser{The selection criteria prioritized models with parameter sizes ranging from 1B to 32B to ensure compatibility with commodity single-GPU machines.}{}
Since in-context learning inherently increases the number of tokens to each prompt (particularly in the many-shot scenario), we opted for a lower cutoff of $8,192$ tokens in the context length. 
This ensures that all selected models could process the longest prompts in our experiments without truncation.

To reduce the memory footprint and computational requirements of our evaluation, we utilize the 8-bit (Q8) quantized versions of the models. This approach significantly reduces memory usage without compromising performance. 
% \textit{Quantization} involves converting model parameters from 16-bit floating-point precision to 8-bit integers, enabling more efficient computation while largely preserving expressive power and accuracy~\cite{jm3}.
\emph{Quantization} involves converting model parameters from higher numerical precisions (e.g.~16-bit floating-point) to formats of lower numerical precision (e.g.~8-bit integers). This process enables more efficient computation and reduced memory usage while largely preserving model performance and expressive power~\citep{jm3}.

% Table~\ref{tab:models} summarizes the selected models. A detailed view along with the HuggingFace links for each model is available at Appendix~\ref{sec:appendix-c-models}.
% Table~\ref{tab:models} \ser{provides an overview of}{summarizes} the selected models, \ser{including their}{listing} context length, \ser{}{and} number of parameters\ser{, and HuggingFace links for easy access}{}.

The model variations used for our evaluation were the following:
%(a more detailed view is available in Appendix~\ref{sec:appendix-c-models}):

\begin{itemize}[nosep, topsep=1pt]
    \item Llama 3 \& 3.1 (8B), Llama 3.2 (1B, 3B)
    \item Mistral Nemo (12B)
    \item Phi 3 Medium (14B), Phi 3.5 Mini (3.8B)
    \item Gemma 2 (2B, 9B, 27B)
    \item Qwen 2 (7B), Qwen 2.5 (14B)
\end{itemize}

\begin{figure*}[t]
    \centering
    \hspace*{-2.5cm}
    \includegraphics[width=1.3\linewidth]{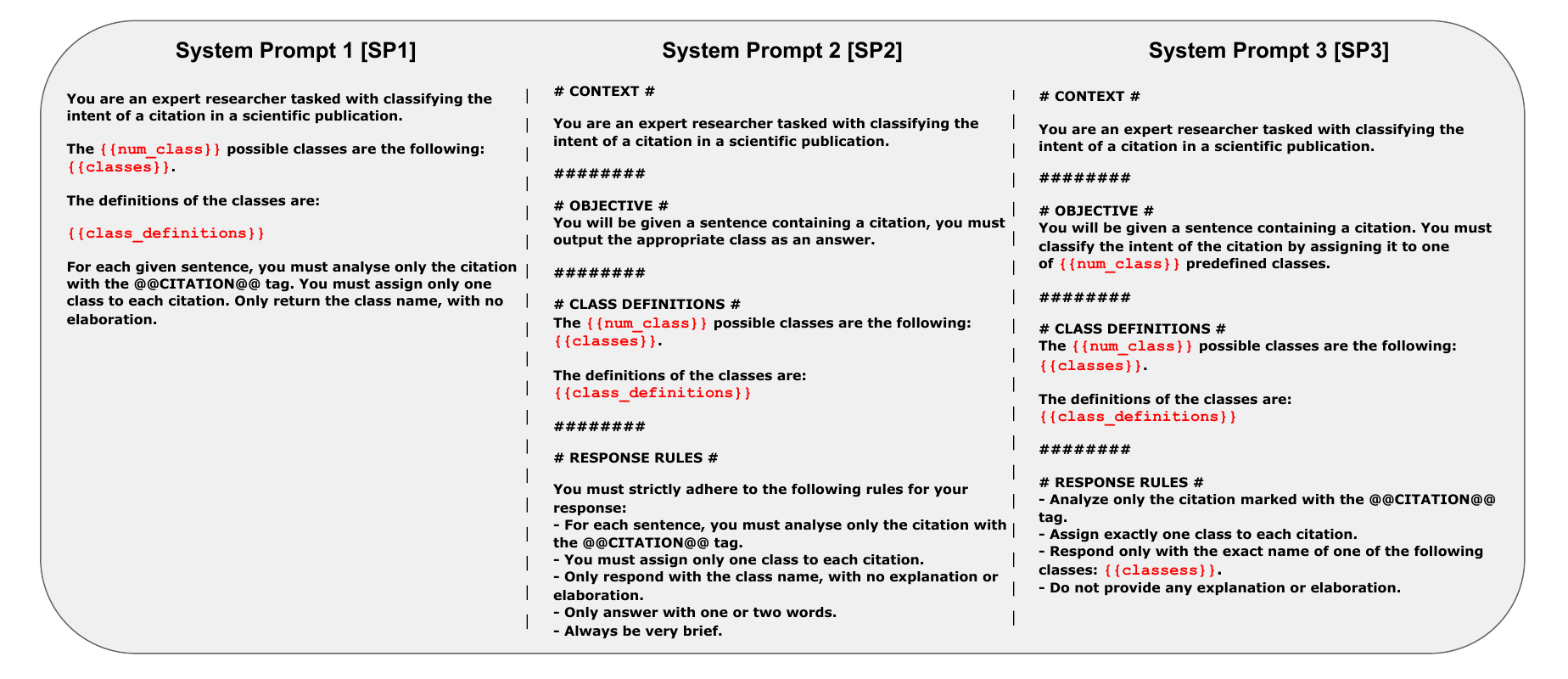}
    \caption{Our System Prompts.}\label{fig:system_prompts}
\end{figure*}

\subsection{Datasets}\label{sec:datasets}
For our experiments, we used two datasets:

\begin{itemize}[nosep, topsep=1pt]
    \item \textit{SciCite}~\citep{CAZ2019} consists of $11,021$ citation strings annotated with three classes: \texttt{Background Information}, \texttt{Method}, and \texttt{Results Comparison}. 
    % \ser{It is widely used in the literature as a benchmark for citation intent tasks.}{}
    \item \textit{ACL-ARC}~\citep{JKH2018} contains $1,941$ citation strings split into six categories: \newline \texttt{Background}, \texttt{Motivation}, \texttt{Uses}, \texttt{Extends}, \texttt{Compares or Contrasts}, and \texttt{Future}. 
    % \ser{It provides a more granular classification of citation intents.}{}
\end{itemize}

These two datasets are widely used in citation intent classification research. 
The SciCite dataset is larger and more diverse, with broad coverage across multiple scientific fields, namely, Computer Science, Medicine, Neuroscience, and Biochemistry, while the ACL-ARC dataset offers a more granular classification scheme, focused on Computational Linguistics. Accross our evaluation, we used the original splits of the datasets, which include training, validation, and test sets (75\% - 8\% - 17\% for SciCite, and 87\% - 7\% - 6\% for ACL-ARC). 

% These characteristics allow us to evaluate model performance across a range of citation intent classification tasks.

% \emph{\color{red} \textbf{Paris:} Will add more details about the datasets from their respective papers.}

\subsection{Configuration Parameters}
\label{sec:configuration-params}
In this section, we describe the configuration parameters of the examined models, including in-context learning methods, system prompts, query templates, example methods, and temperature settings.
% Details on the range of values for these parameters can be found in Table~\ref{tab:experimental-params}.

% \begin{table}[t]
%     \centering
%     \footnotesize
%     \begin{tabular}{cc}
%       \hline
%       \textbf{Parameter} & \textbf{Range of values} \\
%       \midrule
%       \multirow{2}{*}{\textbf{Prompting methods}} & Zero-shot, One-shot, \\
%                                                   & Few-shot, Many-shot \\
%       \midrule
%       \textbf{System prompts} & SP1, SP2, SP3 \\
%       \midrule
%         \textbf{Query templates} & Simple, Multiple-choice \\
%         \textbf{Example methods} & Inline, Roles \\
%         \textbf{Temperature} & $0.0$, $0.2$, $0.5$, $1.0$ \\
%       \hline
%     \end{tabular}
%     \caption{Configuration parameters. \emph{\color{red} (\textbf{Paris:} Might remove)}}\label{tab:experimental-params}
% \end{table}

\subsubsection{In-context Learning Methods (ILM).}\label{sec:in_context_learning}
To evaluate model performance, we applied four prompting methods\footnote{Throughout the paper, we use ``In-context Learning Methods'' and ``Prompting Methods'' interchangibly.} by following the In-Context Learning paradigm \citep{BMR2020}:
% \begin{itemize} [nosep,topsep=1pt]
%     \item \textbf{Zero-shot prompting:} No examples are provided. The model infers the task solely from the system prompt.
%     \item \textbf{One-shot prompting:} One example per class is provided to guide the model.
%     \item \textbf{Few-shot prompting:} Five examples per class are included to offer additional context.
%     \item \textbf{Many-shot prompting:} Ten examples per class are provided, representing a more extensive context.
% \end{itemize}
\textit{Zero-shot} (no examples),
\textit{One-shot} (a single example per class),
\textit{Few-shot} ($5$ examples per class), and
\textit{Many-shot} ($10$ examples per class).

These prompting methods were selected to evaluate whether model performance improves as the number of examples increases, and to identify any saturation point where additional examples yield diminishing or negative returns. 
This aligns with prior literature on in-context learning methods, where these specific configurations have been extensively studied~\citep{icl-survey}.
% \ser{In-context learning is particularly suited to citation intent classification, as it allows the models to leverage their pre-trained contextual understanding without requiring task-specific fine-tuning.}{}
Incorporating examples directly into the prompts allows models to better understand the task and the expected output format, which is especially important for citation intent classification, where the citation context is key to determining the correct label.

\subsubsection{System Prompts (SP).}\label{sec:system_prompts}
System prompts (SP) are foundational to our experimental design, establishing the context and expected behavior for the models in the citation intent classification task. They provide task-specific instructions, class definitions, and output guidelines. We developed and evaluated three distinct system prompt variations, shown in Figure~\ref{fig:system_prompts}.
\begin{itemize}[nosep, topsep=1pt]
    \item \textbf{SP1:} This initial prompt offered a straightforward, intuitive instruction for the classification task.
    \item \textbf{SP2:} This prompt introduced more structure, drawing inspiration from the CO-STAR framework~\citep{costar}. We adapted relevant CO-STAR components by definint the task \emph{Context}, the \emph{Objective} of the model, and detailed \emph{Response Rules}. Given that our task requires a single, specific class label output, the stylistic components of CO-STAR (\emph{Style}, \emph{Tone}, and \emph{Audience}) were not incorporated.
    \item \textbf{SP3:} Building upon SP2, this prompt further refined wording of the \emph{Objective} and \emph{Response Rules}. It also explicitly reiterated the expected labels in the \emph{Response Rules} section, aiming to improve task clarity and consistency in the model's outputs.
\end{itemize}

\subsubsection{Query Templates (QT).}\label{sec:query_templates}
Query templates (QT) define the specific format used to present citation sentences to the model, whether as part of an in-context learning example or as the actual query requiring classification. We evaluated two template structures:
\begin{itemize}[nosep, topsep=1pt]
    \item \textbf{Simple Query:} Initially, we used a basic format where the citation sentence was followed by ``Class:''. For examples, the correct class was appended; for queries, this was left for the model to predict.
    \item \textbf{Multiple-choice Query:} We observed that some models struggled to adhere to the strict class-label-only output expected with the Simple Query, resulting in inconsistent performance. To mitigate this, we introduced a more structured template. Here, after the citation sentence, the model is explicitly prompted to choose the most appropriate citation intent from a presented list of all possible class labels.
\end{itemize}
Although the Multiple-choice template increases the token count per instance, it yielded significant performance improvements (detailed in Section~\ref{sec:parameter_performance}) by providing clearer guidance to the models.
% Query templates define the structure in which citation sentences are presented to the model, both as examples during prompting and as queries during evaluation. 
% Initially, we adopted a \textit{Simple Query} template: 
% the citation sentence was followed by ``Class:'', 
% either pre-filled with the correct class (in examples) or left blank (in queries). However, we observed that some models struggled to align with the expectation of responding solely with the class labels, leading to inconsistent performance and increased variance.
% To address this, we introduced a \textit{Multiple-choice Query} template, where, following the citation sentence, the model is explicitly asked to identify the most likely citation intent, with the possible class labels presented as multiple-choice options. 
% While this approach increases the token count per query, it demonstrates significant performance gains, as discussed in Section~\ref{sec:parameter_performance}.

% \ser{In both templates, minimal post-processing was necessary to extract the predicted class label from the model’s output and ensure alignment with the evaluation criteria.}{}

% \begin{figure}[t]
%     \centering
%     % \hspace*{-4cm}
%     \includegraphics[width=\linewidth]{figures/placeholder_epf_qt.pdf}
%     \caption{The different Example Presentation Formats and Query Templates.}\label{fig:epf_qt}
% \end{figure}

\subsubsection{Example Presentation Formats (EPF).}\label{sec:example_method_variations}

In our experiments, we explored two formats for presenting in-context examples to the models: \emph{Inline} and \emph{Conversational}.

\begin{itemize}[nosep, topsep=1pt]
    \item \textbf{Inline:} Example citation sentences and their corresponding classes (structured according to the defined query templates) are embedded directly and sequentially within the main prompt, immediately preceding the actual query sentence that the model must classify.
    \item \textbf{Conversational:} Examples are structured to simulate a dialogue. Each example consists of a ``user'' turn providing a citation sentence and an ``assistant'' turn providing the correct class. This series of example turns is provided before the final ``user'' query, which presents the new citation sentence for the model to classify. This format is also commonly referred to as a conversational or turn-taking format.
\end{itemize}

% In the Inline approach, example citation sentences and their corresponding classes (as per the defined query templates) are provided directly within the system prompt. After this setup, the prompt then transitions to the evaluation phase, where the ``user'' provides a citation sentence without a corresponding class and the model, acting as the ``assistant'', predicts the correct class.

% The Roles method, on the other hand, simulates a conversational exchange between the ``user'' and ``assistant''.
% Each example follows a back-and-forth interaction, with the ``user'' providing a citation sentence and the ``assistant'' responding with the correct class. 
% After several such examples, the interaction moves to the evaluation phase, where the ``user'' presents a new citation sentence, and the model predicts its class.

% \ser{We did not identify any immediate theoretical advantage to either method, so both were tested to evaluate their potential impact on model performance. Further discussion on the outcomes and insights derived from these methods can be found in Section~\ref{sec:parameter_performance}.}{}

\subsubsection{Temperature (T).}\label{sec:temperature}

Temperature is a hyperparameter that controls the randomness or creativity of a language model's outputs by adjusting the probability distribution of possible next tokens~\citep{jm3}. Lower temperatures (i.e., close to $0$) correspond to greedy decoding, where the model deterministically selects the most probable token at each step, while higher temperatures (closer to $1$) introduce greater variability by allowing the model to sample from a broader range of options.

For classification tasks, we aim for the model to output the most probable class label. 
Therefore, we use a temperature of $0$ as the baseline, ensuring fully deterministic predictions. 
To explore how controlled randomness affects classification performance, we also evaluate higher temperatures, such as $0.2$, $0.5$, and $1.0$.
A temperature of $0.2$ introduces a small degree of randomness but still heavily favors the most probable answer, while $0.5$ strikes a balance between randomness and determinism. At $1.0$, randomness is maximized, allowing us to assess whether excessive stochasticity degrades performance.

\subsection{Technical Specifications}\label{sec:technical_specifications}

We conducted our experiments on an M1 Max Mac Studio with $64$GB of memory, chosen to demonstrate the feasibility of running inference for Citation Intent Classification on commodity hardware.

For model hosting, 
we used LM Studio\footnote{\url{https://lmstudio.ai/}} which offers an intuitive interface for testing and interacting with models hosted on HuggingFace or locally. 
It also supports a local server mode compatible with the OpenAI API, allowing interaction through an API accessible in multiple programming languages. 
A command-line interface (CLI) tool\footnote{\url{https://github.com/lmstudio-ai/lms}} is also available for managing the server without using the UI.

% For hosting the models, we used LM Studio. LM Studio provides:
% \begin{enumerate}
%     \item User Interface (UI) for testing and interacting with models hosted on HuggingFace or locally.
%     \item Local server mode that provides API compatibility with the OpenAI API and supports multiple programming languages.
%     \item Command-line interface (CLI) tool (lms) for managing the server without relying on the UI.
% \end{enumerate}

% While alternatives like Ollama or HuggingFace's transformers library exist, 
% we chose LM Studio for its ease of use and compatibility with the OpenAI API, which aligned with our workflow.

% \ser{Since we are open-sourcing our evaluation platform, \ser{and the models and experimental setups are}{with} configurable \ser{}{models and setups}, 
% we \ser{are considering adding}{plan to add} support for alternative hosting tools in the future to accommodate different user needs and computational environments.}{}

% \emph{ \color{red} \textbf{Paris:} I want to say more about the evaluation framework and technical stuff, so I might make this section smaller and add more details to an appendix called ``Evaluation Platform''.}

\section{Experimental Results and Analysis}\label{sec:results}
This section presents the results of our experiments, focusing on model performance across various configurations and parameter settings. An initial overview of the peak F1-score achieved by each of the twelve models on both the SciCite and ACL-ARC datasets is summarized in Tables~\ref{tab:best-f1-per-model-scicite} and~\ref{tab:best-f1-per-model-aclarc}. The subsections that follow provide a detailed analysis of top-performing model configurations, evaluate overall model rankings, and explore the influence of specific parameters.

\begin{table}[t]
    \caption{Highest F1-Score Performance by Model on SciCite.}\label{tab:best-f1-per-model-scicite}
    \centering
    % \setlength{\tabcolsep}{5pt} % Adjust horizontal padding
    % \scriptsize
    \footnotesize
    \renewcommand{\arraystretch}{1.1}
    \begin{tabular}{clc}
      % \hline
      % \multicolumn{3}{c||}{\textbf{SciCite}} & \multicolumn{3}{c}{\textbf{ACL-ARC}} \\
      \hline
      \textbf{Rank} & \textbf{Model}          & \textbf{F1-Score} \\
      \hline
      1            & Qwen 2.5~--~14B          & 78.33             \\
      2            & Gemma 2~--~27B           & 77.86             \\
      3            & Mistral Nemo~--~12B      & 77.39             \\
      4            & Gemma 2~--~9B            & 75.12             \\
      5            & Phi 3 Medium~--~14B      & 74.67             \\
      6            & LLaMA 3~--~8B            & 74.39             \\
      7            & Qwen 2~--~7B             & 72.89             \\
      8            & LLaMA 3.1~--~8B          & 72.46             \\
      9            & Gemma 2~-~2B             & 68.79             \\
      10           & Phi 3.5 Mini~--~3.8B     & 68.25             \\
      11           & LLaMA 3.2~-~3B           & 67.99             \\
      12           & LLaMA 3.2~-~1B           & 45.44             \\
      \hline
    \end{tabular}
\end{table}

% \emph{\color{red} \textbf{Paris:} \textbf{1}~--~My issue with this section is the is that we don't really show the actual results anywhere (apart from Table \ref{tab:optimal_configurations} which has some cherry-picked results - might get removed for different reasons, see section 4.3 notes). I plan to put the Top-20 or Top-50 in the appendix, but maybe we should put a Top-5 here and reference it in the 4.1 introductory section (it would look like Table \ref{tab:top_5_configurations})? \textbf{2}~--~I also had a section for the statistical tests as a second form of parameter performance analysis (section 4.2.2), but I removed it because it took too much space and didn't really provide any new insights (it was mostly a confirmation of the quantile-based analysis). It is still commented-out in the tex file. We could put it in the appendix, but I'm not sure if it's worth it~--~it also feels that we might be abusing the unlimited appendix at this point.}

\subsection{Model Performance over Configurations}\label{sec:best_models}

Our first evaluation focuses on identifying the top-performing models across the configurations described in Section~\ref{sec:experimental_setup}. We employed two complementary approaches: (i) the Best-Performer Evaluation identifies the single best model, and (ii) the Ranked Evaluation considers the relative performance of all models across configurations.

\subsubsection{Best-Performer Evaluation.}\label{sec:best_performer_evaluation}

To identify the most effective models across configurations, we initially conducted a Best-Performer Evaluation. In particular, for each configuration, we examined a metrics table containing precision, recall, F1-score, and accuracy for all models. 
The model with the highest F1-score\footnote{All F1-scores reported in this paper are macro-averaged F1-scores (macro-F1).} in each configuration was selected as the best-performing model~--~in case of a tie, we used Accuracy as the deciding factor. 
% We then aggregated the results across all configurations and grouped them by model to determine how often each model achieved the best performance accross our configurations. \ser{This approach allowed us to quantify the dominance of each model across the experimental space.}{}

\begin{table}[t]
    \caption{Highest F1-Score Performance by Model on ACL-ARC.}\label{tab:best-f1-per-model-aclarc}
    \centering
    % \setlength{\tabcolsep}{5pt} % Adjust horizontal padding
    % \scriptsize
    \footnotesize
    \renewcommand{\arraystretch}{1.1}
    \begin{tabular}{clc}
      % \hline
      % \multicolumn{3}{c||}{\textbf{SciCite}} & \multicolumn{3}{c}{\textbf{ACL-ARC}} \\
      \hline
      \textbf{Rank} & \textbf{Model}          & \textbf{F1-Score} \\
      \hline
      1             & Qwen 2.5~--~14B         & 63.68         \\
      2             & Gemma 2~--~27B          & 58.95         \\
      3             & Gemma 2~--~9B           & 57.19         \\
      4             & Qwen 2~--~7B            & 51.26         \\
      5             & LLaMA 3.1~--~8B         & 48.45         \\
      6             & Mistral Nemo~--~12B     & 48.11         \\
      7             & Phi 3.5 Mini~--~3.8B    & 43.74         \\
      8             & Phi 3 Medium~-~14B      & 43.46         \\
      9             & Gemma 2~-~2B            & 40.96         \\
      10            & LLaMA 3.2~-~3B          & 40.07         \\
      11            & LLaMA 3~-~8B            & 38.06         \\
      12            & LLaMA 3.2~-~1B          & 24.60         \\
      \hline
    \end{tabular}
\end{table}

\begin{table*}[t]
    \caption{Model Ranking based on Best-Performing Count.}\label{tab:best_performer_evaluation}
    \centering
    % \footnotesize
    \renewcommand{\arraystretch}{1.1}
    \setlength{\tabcolsep}{5pt} % Adjust horizontal padding
    % \hspace*{-1.85cm}
    % \hspace*{-1cm}
    % \scriptsize
    \footnotesize
    \begin{tabular}{cclcccccc}
      \hline
      \textbf{Dataset}            &  \textbf{Rank} &   \textbf{Model}          & \textbf{Overall}    & \textbf{Zero-Shot}  & \textbf{One-Shot} & \textbf{Few-Shot}  & \textbf{Many-Shot} \\
    %   \multirow{2}{*}{\textbf{Dataset}}            &  \multirow{2}{*}{\textbf{Rank}} &   \multirow{2}{*}{\textbf{Model}}          & \multirow{2}{*}{\textbf{Overall}}    & \textbf{Zero-}  & \textbf{One-} & \textbf{Few-}  & \textbf{Many-} \\
    %                               &                &                           &                     & \textbf{Shot}   & \textbf{Shot} & \textbf{Shot}  & \textbf{Shot} \\
      \hline
      \multirow{4}{*}{SciCite}  & 1             &   Qwen 2.5~--~14B         & 125                               & 24                  &  42               & 28                 & 31                 \\
                                & 2             &   Mistral Nemo~--~12B     & 25                                & 0                   &  0                & 14                 & 11                 \\
                                & 4             &   Gemma 2~--~27B          & 10                                 & 0                   &  0                & 4                  & 6                  \\
                                & 4             &   Gemma 2~--~9B           & 8                                 & 0                   &  6                & 2                  & 0                  \\
      \hline
      \multirow{3}{*}{ACL-ARC}  & 1             &   Qwen 2.5~--~14B         & 129                          & 19                   &  22                & 40                  & 28                  \\
                                & 2             &   Gemma 2~--~27B          & 28                           & 0                    &  21                & 7                  & 0                  \\
                                & 3             &   Gemma 2~--~9B           & 11                           & 5                    &  5                & 1                  & 0                  \\
      \hline
    \end{tabular}
\end{table*}

The aggregated results of our evaluation are presented in Table~\ref{tab:best_performer_evaluation}.
%additionally, Table~\ref{tab:top_5_configurations} lists the top-$5$ best-performing configurations on each dataset.
It is evident that Qwen 2.5 14B was the most dominant model on both datasets, significantly outperforming others across all prompting methods.

\subsubsection{Ranked Evaluation}\label{sec:ranked_evaluation}

While the Best-Performer Evaluation provided insights into the most dominant models across configurations, it lacked the ability to capture nuances among high-performing models. For instance, a model that consistently ranked second in multiple configurations would not be reflected in that approach.
To address this limitation, we adopted a ranking methodology inspired by Reciprocal Rank Fusion (RRF)~\citep{cormack2009reciprocal}. 
This approach allowed us to evaluate all $12$ models considering their relative performance across different configurations.

For each experimental configuration \(c \in C\),  models \(M =\{ m_1, m_2, \dots, m_{12} \}\) were ranked based on their F1-scores~--~in case of a tie, Accuracy was used to determine the rank.
Each model was then assigned a score based on its rank, where the score \(S(m_{k}, c)\) for the \(k\)-th ranked model in configuration \(c\) was defined as \(S(m_{k}, c) = \frac{1}{k}\). 
This score is inversely proportional to the rank, meaning that higher-ranked models (lower-\(k\)) receive higher scores, while lower-ranked models (higher-\(k\)) receive lower scores. 
To calculate the overall ranking score for each model \(m\), we aggregated its scores across all configurations \(c \in C\):
\begin{equation}
    RankingScore(m) = \sum_{c \in C} S(m_{k}, c)
\end{equation}
% The results of this ranked evaluation are presented in 
Tables~\ref{tab:rrf_evaluation_scicite} and~\ref{tab:rrf_evaluation_aclarc} present the results of this ranked evaluation. The ranked evaluation aligns with the Best-Performer Evaluation in identifying Qwen 2.5 14B as the most dominant model. This view also allows us to gain insights into the relative performance of other the models, 
highlighting distinctions among high performers, as well as the performance of lower-ranked models.
% \ser{This additional layer of analysis offers a more nuanced understanding of model performance across the experimental space.}{}

\subsection{Parameter Performance Analysis}\label{sec:parameter_performance}

In this section, we examine how different parameter configurations drive model performance. Identifying the most impactful parameters was challenging due to the extensive search space ($168$ configurations per dataset, totaling $3,841$ experiments for all models). To address this, we conducted a Quantile-Based Analysis focusing on the top $5$\% of configurations based on F1-scores. Examining parameter distributions within this high-performing subset (Table~\ref{tab:parameter_performance}) reveals trends for optimal configurations across both datasets.

\begin{table}[t]
    \centering
    \footnotesize
    \renewcommand{\arraystretch}{1.1}
    \begin{tabular}{clc}
      \hline
      \textbf{Rank}   &   \textbf{Model}          &   \textbf{Ranking Score} \\
      \hline
      1	            &   Qwen 2.5~--~14B         &   144.163        \\
      2	            &   Mistral Nemo~--~12B     &   68.479         \\
      3	            &   Gemma 2~--~27B          &   63.369         \\
      4	            &   Gemma 2~--~9B           &   59.760         \\
      5	            &   Qwen 2~--~7B            &   38.617         \\
      6	            &   LLaMA 3.1~--~8B         &   27.624         \\
      7	            &   LLaMA 3~--~8B           &   26.159         \\
      8	            &   Phi 3.5 Mini~--~3.8B    &   22.088         \\
      9	            &   Phi 3 Medium~--~14B     &   19.370         \\
      10	            &   Gemma 2~-~2B            &   19.253         \\
      11              &   LLaMA 3.2~-~3B          &   18.286         \\
      12              &   LLaMA 3.2~-~1B          &   14.136         \\
      \hline
    \end{tabular}
    \caption{Model ranking based on RRF on SciCite.}\label{tab:rrf_evaluation_scicite}
\end{table}

% For this analysis, we focused exclusively on the SciCite dataset, and the insights gained from it were then used to limit the range of experiments conducted on the ACL-ARC dataset.
% \ser{The}{A detailed} analysis of the parameter performance on the ACL-ARC dataset is available in \ser{the supplementary material}{\color{red}Appendix XXX}. 
% The findings from this analysis can be seen in Table~\ref{tab:parameter_performance} and are summarized below.

% For the \emph{In-context Learning Method}, we discern that Few-shot prompting was the most common approach among the top-performing configurations on SciCite, accounting for 46.53\% of these results. Many-shot prompting followed with 35.64\%, while One-shot and Zero-shot prompting were significantly less represented at 13.86\% and 3.96\%, respectively. A similar, even more pronounced trend was observed on the ACL-ARC dataset, where Few-shot prompting was dominant, appearing in 61.11\% of top configurations. Many-shot (23.33\%) and One-shot (15.55\%) methods were less frequent in this subset for ACL-ARC, and Zero-shot configurations did not feature among its top 5\%. This suggests that providing a moderate number of examples generally leads to better performance, particularly five examples per class in the Few-shot setting, although a larger number of examples (Many-shot) also frequently contributes to high performance on SciCite.

For the \emph{In-context Learning Method}, Few-shot prompting was most prevalent in top SciCite configurations ($46.53$\%), with Many-shot also prominent ($35.64$\%). This Few-shot preference was stronger on ACL-ARC ($61.11$\%), where Many-shot ($23.33$\%) and One-shot ($15.55$\%) were less frequent, and Zero-shot was absent from the top 5\%. These results suggest moderate example counts (Few-shot) are broadly effective, though Many-shot also performs well on SciCite.

The optimal \emph{Temperature} settings showed divergence between the datasets. SciCite favored lower values, with T=0.0 ($30.69$\%) and T=0.2 ($29.70$\%) being most frequent, indicating a preference for deterministic outputs, while higher temperatures (T=0.5, T=1.0, both $19.80$\%) were less common. In contrast, ACL-ARC displayed a more balanced distribution, where T=0.0 and T=1.0 (both $25.55$\%) were slightly ahead of T=0.2 and T=0.5 (both $24.44$\%), suggesting that varied levels of randomness can yield top results.

% Turning to the \emph{Temperature} parameter, the optimal settings showed more divergence between the datasets. On SciCite, lower temperatures were clearly favored, with T=0.0 (30.69\%) and T=0.2 (29.70\%) being the most frequent settings in the top configurations, suggesting that more deterministic outputs are generally beneficial. Higher temperatures, such as T=0.5 and T=1.0 (both at 19.80\%), were less common. For the ACL-ARC dataset, however, the distribution among temperatures was more balanced; T=0.0 and T=1.0 (both 25.55\%) were only marginally more frequent than T=0.2 and T=0.5 (both 24.44\%), indicating that both deterministic outputs and those with some degree of randomness can lead to top-tier results.

% The analysis of the \emph{System Prompt} parameter revealed a common preference for SP3 across both datasets, which was present in 44.55\% of top SciCite configurations and 42.22\% of top ACL-ARC configurations. Beyond this leading choice, secondary preferences varied: for SciCite, SP1 and SP2 followed with equal representation (27.72\% each), whereas on ACL-ARC, SP1 (31.11\%) was somewhat more prevalent in the top configurations than SP2 (26.66\%). These results highlight a general advantage for the more structured and explicit SP3, even if simpler prompts also contribute to high-performing configurations.

A shared preference for \emph{System Prompt} SP3 emerged across both datasets, found in $44.55$\% of top SciCite and $42.22$\% of top ACL-ARC configurations. Secondary preferences, however, varied. SciCite showed equal representation for SP1 and SP2 (both $27.72$\%), whereas ACL-ARC favored SP1 ($31.11$\%) over SP2 ($26.66$\%). This indicates an advantage for the structured SP3, though simpler prompts also achieved high performance.

% A particularly decisive trend emerged with the \emph{Query Template} parameter. The Multiple-Choice query template was overwhelmingly favored, appearing in a striking 79.21\% of the top-performing configurations on SciCite and a significant 64.44\% on ACL-ARC. Conversely, the Simple query template was found in only 20.79\% of top SciCite results and 35.55\% of those for ACL-ARC. This strong preference indicates that providing explicit multiple-choice options is considerably more effective for the evaluated tasks, likely by better guiding the models' output format.

The \emph{Query Template} parameter revealed a decisive trend, as the Multiple-Choice template was overwhelmingly favored on both SciCite ($79.21$\%) and ACL-ARC ($64.44$\%) when compared to the Simple template (SciCite: $20.79$\%; ACL-ARC: $35.55$\%). This highlights the significant effectiveness of explicit multiple-choice options.

Finally, considering the \emph{Example Presentation Format}, the Conversational style was more frequent in top configurations for SciCite ($54.64$\%) and especially ACL-ARC ($65.55$\%), over the Inline format (SciCite: $45.36$\%; ACL-ARC: $34.44$\%). This suggests structuring examples conversationally generally yields better results.

% Finally, considering the \emph{Example Presentation Format}, the Conversational format was more frequently associated with top-performing configurations for both SciCite (54.64\%) and, more notably, for ACL-ARC (65.55\%). The Inline format, while still effective, accounted for the remaining 45.36\% on SciCite and 34.44\% on ACL-ARC. This suggests that structuring in-context examples within a simulated dialogue format generally yields better results than a direct inline presentation.

\begin{table}[t]
    \centering
    \footnotesize
    \renewcommand{\arraystretch}{1.1}
    \begin{tabular}{clc}
      \hline
      \textbf{Rank} &   \textbf{Model}          &   \textbf{Ranking Score} \\
      \hline
      1	            &   Qwen 2.5~--~14B         &   146.496         \\
      2	            &   Gemma 2~--~27B          &   62.973          \\
      3	            &   Gemma 2~--~9B           &   55.318          \\
      4	            &   Qwen 2~--~7B            &   41.496          \\
      5	            &   LLaMA 3.1~--~8B         &   40.844          \\
      6	            &   Phi 3.5 Mini~--~3.8B    &   30.470          \\
      7	            &   Mistral Nemo~--~12B     &   30.285          \\
      8	            &   LLaMA 3.2~-~3B          &   23.984          \\
      9               &   Phi 3 Medium~-~14B      &   21.269          \\
      10              &   LLaMA 3.2~-~1B          &   16.796          \\
      11	            &   Gemma 2~-~2B            &   16.534          \\
      12              &   LLaMA 3~-~8B            &   16.359          \\
      \hline
    \end{tabular}
    \caption{Model ranking based on RRF on ACL-ARC.}\label{tab:rrf_evaluation_aclarc}
\end{table}

\begin{table*}[t]
    \caption{Parameter Performance Analysis of Top 5\% Configurations.}\label{tab:parameter_performance}
    \centering
    \setlength{\tabcolsep}{5pt}

    \footnotesize
    \begin{tabular}{l|c|cc||cc}
      \hline
      \multirow{2}{*}{\textbf{Parameter}}                   & \multirow{2}{*}{\textbf{Setting}}      & \multicolumn{2}{c||}{\textbf{SciCite}}  & \multicolumn{2}{c}{\textbf{ACL-ARC}}  \\
      \cline{3-4}\cline{5-6}
                                                            &                                        & \textbf{Count} & \textbf{Percent}      & \textbf{Count} & \textbf{Percent}     \\
      \hline
      \multirow{4}{*}{\textbf{In-context Learning Method}}  & Few-shot                               & 47             & 46.53\%              & 55             & 61.11\%              \\
                                                            & Many-shot                              & 38             & 35.64\%              & 21             & 23.33\%              \\
                                                            & One-shot                               & 14             & 13.86\%              & 14             & 15.55\%              \\
                                                            & Zero-shot                              & 4              & 3.96\%               & -              & -              \\
      \hline
      \multirow{4}{*}{\textbf{Temperature}}                 & 0.0                                    & 31             & 30.69\%              & 23             & 25.55\%              \\
                                                            & 0.2                                    & 30             & 29.70\%              & 22             & 24.44\%              \\
                                                            & 0.5                                    & 20             & 19.80\%              & 22             & 24.44\%              \\ 
                                                            & 1.0                                    & 20             & 19.80\%              & 23             & 25.55\%              \\
      \hline
      \multirow{3}{*}{\textbf{System Prompt}}               & SP3                                    & 45             & 44.55\%              & 38             & 42.22\%              \\
                                                            & SP2                                    & 28             & 27.72\%              & 24             & 26.66\%              \\
                                                            & SP1                                    & 28             & 27.72\%              & 28             & 31.11\%              \\ 
      \hline
      \multirow{2}{*}{\textbf{Query Template}}              & Multiple-Choice                        & 80             & 79.21\%              & 58             & 64.44\%              \\
                                                            & Simple                                 & 21             & 20.79\%              & 32             & 35.55\%              \\
      \hline
      \multirow{2}{*}{\textbf{Example Presentation}}        & Conversational                         & 53             & 54.64\%              & 59             & 65.55\%              \\
                                                            & Inline                                 & 44             & 45.36\%              & 31             & 34.44\%              \\
      \hline
    \end{tabular}
\end{table*}

\begin{table*}[t]
    \caption{Chi-Square Test of Independence results for parameter settings and F1-Scores. \emph{(Significance levels: ***p < 0.001, **p < 0.01, *p < 0.05)}}\label{tab:chi-square-results}
    \centering
    \setlength{\tabcolsep}{5pt}
    \renewcommand{\arraystretch}{1.3}
    \footnotesize
    % \hspace*{-0.7cm}
    \begin{tabular}{l|cl||cl}
        \hline
        \multirow{2}{*}{\textbf{Parameter}}     & \multicolumn{2}{c||}{\textbf{SciCite}}  & \multicolumn{2}{c}{\textbf{ACL-ARC}}  \\
        \cline{2-3}\cline{4-5}
        & \textbf{Chi-Square ($\chi^2$)} & \textbf{p-value} & \textbf{Chi-Square ($\chi^2$)} & \textbf{p-value} \\
        \hline
        \textbf{In-context Learning Method}     & 29.857  & 1.48e-06 *** & 49.025  & 1.29e-10 *** \\
        \textbf{Temperature}                    & 4.617   & 0.202        & 0.047   & 0.9973        \\
        \textbf{System Prompt}                  & 6.024   & 0.0491 *       & 3.647   & 0.1615        \\
        \textbf{Query Template}                 & 35.063  & 3.19e-09 *** & 7.305   & 0.0068 ** \\
        \textbf{Example Presentation}           & 0.699   & 0.4031        & 8.604   & 0.0033 ** \\
        \hline
    \end{tabular}
\end{table*}

These findings offer valuable insights into the parameter settings that consistently enhance performance across the two datasets.

% We discern that few-shot prompting was the most common prompting method among the top-performing configurations, accounting for nearly 45\% of the results. 
% Many-shot prompting followed closely with 34.83\%, while one-shot and zero-shot prompting were significantly less represented, with only 16.85\% and 3.37\%, respectively. 
% This suggests that providing several examples during prompting generally leads to better performance, with the optimal achieved at five examples (few-shot), followed by a slight decrease at larger numbers of examples.

% In contrast, the temperature parameter exhibited a relatively even distribution among the top-performing configurations, with $0.0$ and $0.2$ being the most frequent settings (30.34\% each). Higher temperatures, such as $1.0$ and $0.5$, 
% were less common, suggesting that lower temperatures contribute to more consistent and deterministic outputs, 
% which align better with the task requirements.

% Similarly, the system prompt parameter was fairly evenly distributed among the top configurations, with SP3 being slightly more frequent (35.96\%), followed by SP2 (32.58\%), and SP1 (31.46\%). 
% While all system prompts contributed to high performance, 
% these results suggest that SP3 has offered a slight advantage in effectiveness.

\subsubsection{Chi-Square Statistical Test}\label{sec:chi_square_test}

To complement our parameter performance analysis, we also conducted a \emph{Chi-Square Test of Independence}~\citep{chi-square} to evaluate the relationship between parameter settings and F1-scores. This approach allowed us to determine whether specific parameter settings were significantly associated with performance outcomes and validate the results of our primary analysis. The results of the test (summarized in Table~\ref{tab:chi-square-results}) largely reinforces the findings from our quantile-based analysis regarding the influence of different parameter settings.

% The test revealed several significant relationships between parameter settings and performance outcomes. Specifically, the \emph{In-context Learning Method} ($\chi^2 = 22.46$, $p = 5.23e-05$), \emph{System Prompt} ($\chi^2 = 14.97$, $p = 0.00056$), and \emph{Query Template} ($\chi^2 = 41.86$, $p = 9.79e-11$) were all found to have statistically significant associations with performance, indicating that these parameters play a critical role in influencing F1-scores. In contrast, \emph{Temperature} ($\chi^2 = 4.31$, $p = 0.230$) and \emph{Example Presentation Format} ($\chi^2 = 0.01$, $p = 0.911$) did not show significant relationships, suggesting that these parameters may not independently affect performance outcomes. These results are aligned with the findings of our Quantile-based analysis.

The \emph{In-context Learning Method} showed a highly significant association with F1-scores for both SciCite ($p < 0.001$) and ACL-ARC ($p < 0.001$). This statistically underscores the importance of the number of examples provided, aligning with our earlier observation that Few-shot and Many-shot configurations were prevalent among the top performers.

Similarly, the \emph{Query Template} was found to be highly significant for SciCite ($p < 0.001$) and also significant for ACL-ARC ($p < 0.01$). This supports the strong dominance of the Multiple-Choice template in the top 5\% configurations, highlighting its critical role in guiding model output effectively.

Other parameters presented more nuanced relationships. The \emph{System Prompt} was statistically significant for SciCite ($p < 0.05$), corroborating the advantage seen for SP3 in its quantile analysis. However, for ACL-ARC, while SP3 was also frequent in top configurations, the choice of system prompt did not emerge as a statistically significant factor overall.

Conversely, the \emph{Example Presentation Format} was not statistically significant for SciCite, which aligns with the relatively balanced distribution (Conversational at 54.64\%, Inline at 45.36\%) observed in its top configurations. For ACL-ARC, however, this parameter was significant ($p < 0.01$), providing statistical backing for the clearer preference (65.55\%) for the Conversational method seen in its quantile analysis.

\begin{table*}[t]
    \caption{Training Parameters for Fine-Tuning Qwen 2.5~--~14B.}\label{tab:training_parameters}
    \centering
    % \scriptsize
    \footnotesize
    \setlength{\tabcolsep}{5pt}
    \renewcommand{\arraystretch}{1.1}
    % \hspace*{-0.55cm}
    \begin{tabular}{lcl}
        \hline
        \textbf{Parameter}  & \textbf{Value}    & \textbf{Explanation} \\
        \hline
        Learning Rate       & 5e-5              & Controls the step size for updating model weights during training. \\
        Epochs              & 10                & Number of complete passes through the entire training dataset. \\
        Batch Size          & 16                & Number of training examples processed in one iteration. \\
        Cutoff Length       & 512               & Maximum sequence length (in tokens) for model inputs. \\
        Optimizer           & AdamW             & Algorithm used to adjust model weights to minimize loss. \\
        Compute Type        & fp16              & Numerical precision used for computations. \\
        Warmup Steps        & 500               & Number of initial steps where the learning rate gradually increases. \\
        DeepSpeed Offload   & Enabled           & Moves optimizer states/gradients to CPU RAM to save GPU memory. \\
        DeepSpeed Stage     & 3                 & Level of DeepSpeed ZeRO optimization. \\
        LoRA Rank           & 8                 & Dimension of the trainable low-rank matrices added by LoRA. \\
        LoRA Alpha          & 16                & Scaling factor for the LoRA updates relative to the rank. \\
        LoRA Dropout        & 0.1               & Dropout probability applied to the LoRA layers for regularization. \\
        \hline
    \end{tabular}
\end{table*}

Finally, the \emph{Temperature} parameter did not show a statistically significant relationship with F1-scores on either dataset. This finding is consistent with the quantile analysis where, for SciCite, multiple temperature settings were present in top configurations (though lower ones were more frequent), and for ACL-ARC, the distribution was particularly balanced, suggesting that temperature, within the tested ranges, was not as decisive a factor as other parameters.

\section{Fine-tuning}\label{sec:fine_tuning}

In this section, we investigate the impact of fine-tuning in the the performance of the instruction-tuned Qwen 2.5 14B model on the task of citation intent classification. 
% \ser{While achieving state-of-the-art results was a secondary goal, the}{Our} primary focus \ser{of this study was}{is} to demonstrate the viability of large language models for this task and explore their potential to generalize to domain-specific challenges.

\subsection{Training Configuration}

For this experiment, we used the SciCite and ACL-ARC datasets, which were converted into the Alpaca format~\citep{alpaca}. This format includes a system prompt, an instruction, the citing sentence, and the true label. 
% An example of our training data in the Alpaca format is included in the appendix, with the full training dataset available in the supplementary material. 
The use of Supervised Fine-Tuning (SFT) was motivated by its ability to adapt pre-trained models to specific tasks using minimal labeled data, making it an effective approach for citation intent classification.
To perform the fine-tuning, we used the original training set of each dataset (8,243 examples for SciCite and 1,688 for ACL-ARC). For testing, we used the original test sets as seen in the previous experiments.

To fine-tune the Qwen 2.5 14B Instruct model, we used LLaMA-Factory~\citep{llamafactory} on an AWS g6e.12xlarge EC2 instance equipped with 4 NVIDIA L40S GPUs, providing a combined GPU memory of 192GB. The training process was conducted using fp16 mixed precision to optimize memory usage and speed. To prevent memory issues during fine-tuning, we employed DeepSpeed ZeRO Stage 3 Offload~\citep{deepspeed}, which enabled efficient memory management by offloading optimizer states and gradients to the CPU. The training parameters included a learning rate of 5e-5, a batch size of 16, and 10 epochs. The model was optimized using AdamW, with a cutoff length of 512 tokens for input sequences. 

To optimize the fine-tuning process, we employed Low-Rank Adaptation (LoRA), a widely used parameter-efficient fine-tuning (PEFT) method~\citep{lora,peft-survey}. LoRA enables efficient adaptation of large language models by freezing the pre-trained model weights and introducing trainable low-rank matrices into the Transformer layers, significantly reducing the number of trainable parameters while maintaining performance. For this experiment, we configured LoRA with a rank of 8, alpha of 16, and a dropout of 0.1, which allowed us to fine-tune the model effectively without exceeding memory constraints. The outlined training parameters are also summarized in Table~\ref{tab:training_parameters}.

% \begin{table}[t]
%     \centering
%     \footnotesize
%     \begin{tabular}{lc}
%         \hline
%         \textbf{Parameter}           & \textbf{Value} \\
%         \hline
%         Learning Rate                & 5e-5           \\
%         Epochs                       & 10             \\
%         Batch Size                   & 16             \\
%         Cutoff Length                & 512            \\
%         Optimizer                    & AdamW          \\
%         Compute Type                 & fp16           \\
%         Warmup Steps                 & 500            \\
%         DeepSpeed Offload            & Enabled        \\
%         DeepSpeed Stage              & 3              \\
%         LoRA Rank                    & 8              \\
%         LoRA Alpha                   & 16             \\
%         LoRA Dropout                 & 0.1            \\
%         \hline
%     \end{tabular}
%     \caption{Training Parameters for Fine-Tuning Qwen 2.5 14B.}\label{tab:training_parameters}
% \end{table}

\begin{table}[t]
    \centering
    \footnotesize
    \begin{tabular}{lcc}
        \toprule
        \textbf{Model}      & \textbf{SciCite}  & \textbf{ACL-ARC}  \\
        \midrule
        Instruct Q8         & 78.33              &  63.68             \\
        Instruct FP16       & 80.41              &  65.64             \\
        Fine-tuned Q8        & 86.47             &  \textbf{68.48}    \\
        Fine-tuned FP16      & \textbf{86.84}    &  67.73             \\
        \bottomrule
    \end{tabular}
    \caption{F1-Score Performance of Qwen 2.5~--~14B Instruct and Fine-tuned variants on the SciCite and ACL-ARC datasets.}\label{tab:results-simple}
\end{table}

\subsection{Results}
% {\color{red} \textbf{Paris:} Discuss the increase of performance in both datasets. Maybe add a plot of the changes per prompting method, since there are large differences in performance.}

\begin{table*}[t]
    \caption{F1-Score Performance of Qwen 2.5 – 14B Instruct and Fine-tuned variants, divided by prompting method.}\label{tab:results-pm}
    \centering
    \footnotesize
    \setlength{\tabcolsep}{5pt}
    \renewcommand{\arraystretch}{1.1}
    % \hspace*{-0.3cm}
    \begin{tabular}{clcccc}
      \hline
      \textbf{Dataset}          & \textbf{Model}    & \textbf{Zero-Shot}   & \textbf{One-Shot}   & \textbf{Few-Shot}   & \textbf{Many-Shot}  \\

      \hline
      \multirow{4}{*}{SciCite}  & Instruct Q8       & 75.74             & 76.32             & 78.33             & 78.23             \\
                                & Instruct FP16     & 75.38             & 77.22             & 80.41             & 78.94             \\
                                & Fine-tuned Q8     & \textbf{84.84}   & \textbf{85.46}     & 86.47             & 85.62             \\
                                & Fine-tuned FP16   & 84.49            & 85.38              & \textbf{86.84}    & \textbf{85.79}    \\
      \hline
      \multirow{4}{*}{ACL-ARC}  & Instruct Q8       & 52.29             & 60.02             & 62.86             & 63.68             \\
                                & Instruct FP16     & 52.47             & 61.73             & 65.64             & 64.39             \\
                                & Fine-tuned Q8     & 59.92             & \textbf{68.48}    & \textbf{67.58}    & 67.19             \\
                                & Fine-tuned FP16   & \textbf{60.29}    & 67.73             & 66.94             & \textbf{67.62}    \\
      \hline
    \end{tabular}
\end{table*}

After fine-tuning, we evaluated the new model using the same prompting methods as in Section~\ref{sec:in_context_learning} (i.e., zero-shot, one-shot, few-shot, and many-shot) to draw comparisons with the instruction-tuned Qwen 2.5 14B baseline. While our initial extensive experiments with in-context learning focused on 8-bit quantized models (Q8), for this fine-tuning evaluation, we also assessed the 16-bit floating-point (FP16) version of the model to examine the impact of numerical precision on performance.

The overall results, presented in Table~\ref{tab:results-simple}, clearly show that supervised fine-tuning significantly improved performance on both the SciCite and ACL-ARC datasets. On SciCite, the fine-tuned FP16 model achieved a peak F1-score of $86.84$. Compared to the instruction-tuned FP16 baseline (F1-score of 80.41), this represents an $8.0$\% relative improvement. On ACL-ARC, the fine-tuned Q8 model obtained the highest F1-score of $68.48$. This constitutes a relative improvement of nearly 4.3\% over the instruction-tuned FP16 baseline (F1-score of $65.64$). Overall, the highest F1-score achieved after fine-tuning was $86.84$ on SciCite and $68.48$ on ACL-ARC.

Examining the impact of numerical precision, the fine-tuned FP16 and Q8 models showed minor performance differences. FP16 models generally outperformed Q8 on SciCite, while the fine-tuned Q8 model slightly outperformed its FP16 counterpart on ACL-ARC. This suggests that while FP16 precision can offer a slight advantage, Q8 remains highly competitive and may offer robustness, particularly for tasks with more granular classification schemes like ACL-ARC's 6-class setup.

Table~\ref{tab:results-pm} highlights the impact of fine-tuning across the different prompting methods. On SciCite, fine-tuning led to consistent F1-score improvements across all prompting scenarios. These gains were particularly substantial in zero-shot and one-shot settings; for instance, the fine-tuned Q8 model improved upon the instruction-tuned FP16 baseline by over $12.5$\% in the zero-shot setting ($84.84$ vs $75.38$) and by nearly $10.7$\% in the one-shot setting ($85.46$ vs $77.22$). In the few-shot and many-shot setups for SciCite, performance with the fine-tuned FP16 model approached saturation, achieving the highest F1-scores of $86.84$ and $85.79$, respectively.

Similar trends were observed on the ACL-ARC dataset. The fine-tuned models generally outperformed their instruction-tuned counterparts across prompting methods, with the fine-tuned Q8 model achieving the highest F1-scores in one-shot ($68.48$) and few-shot ($67.58$) setups. In contrast, the fine-tuned FP16 model yielded stronger results in the zero-shot ($60.29$) and many-shot ($67.62$) scenarios.

These results demonstrate that supervised fine-tuning not only boosts overall performance but also particularly enhances generalization in low-context (zero-shot and one-shot) scenarios, while still effectively leveraging the additional context provided in few-shot and many-shot settings.

\subsection{Discussion}

The primary goal of this work was to investigate the capability of large language models to perform citation intent classification, rather than to compete directly with state-of-the-art methods. Nonetheless, it is worth noting that the fine-tuned model achieved performance on SciCite within less than a 3\% margin of the best-reported results in the literature, surpassing most of the existing approaches (see Table~\ref{tab:literature-f1} - Section~\ref{sec:related_work}). This demonstrates that LLMs can approach the performance of specialized models, even without task-specific architectures or optimization.

We consider a key advantage of LLMs to be their ease of use and deployment. Tools such as LM Studio and Ollama\footnote{\url{https://ollama.com/}} allow models like those outlined in this paper to be deployed locally with zero technical expertise, making them accessible to users without a computer science background. 
In addition to their accessibility, LLMs offer significant adaptability. Unlike traditional methods, which often require complex pretraining or domain-specific tuning, LLMs can be fine-tuned for a wide range of scientometric tasks, such as citation recommendation, paper summarization, or trend analysis, without the need for bespoke architectures. Furthermore, LLMs can scale effectively to new scientific domains or datasets, requiring only small amounts of task-specific data to adapt to underexplored scientific fields; our fine-tuned models required only the several thousand citing sentences provided by our task-specific datasets for adaptation. This contrasts sharply with models like SciBERT, whose effectiveness stems from deliberate pre-training exclusively on a large corpus of scientific papers (millions of articles), optimizing it for scientific language understanding. While the general-purpose LLMs used in our study were likely exposed to scientific text within their vast, diverse pre-training data drawn from the web, this exposure is \emph{incidental} rather than \emph{targeted}. They were not specifically pre-trained or architected with the primary goal of processing scientific literature, unlike SciBERT. Our results demonstrate that even without such specialized scientific pre-training, general-purpose LLMs can achieve competitive performance through efficient fine-tuning on minimal task-specific data.

% We consider a key advantage of LLMs to be their ease of use and deployment. Unlike traditional methods, which are often narrowly optimized for specific tasks, LLMs offer a flexible solution that can be fine-tuned for a variety of scientometric tasks with minimal effort. For example, the same fine-tuned LLM used for citation intent classification can also be adapted for tasks such as citation recommendation, paper summarization, or trend analysis in research topics, eliminating the need for bespoke architectures for each task. Furthermore, tools such as LM Studio and Ollama allow models like those outlined in this paper to be deployed locally with zero technical expertise, making them accessible to users with limited technical knowledge.

% We also find that fine-tuned LLMs are particularly well-suited for scalability to new domains or datasets. Traditional methods often require significant retraining or re-engineering to adapt to new datasets or emerging fields of research. In contrast, LLMs can be fine-tuned on a small subset of relevant data, making them particularly effective for analyzing citation intent in underexplored scientific domains This scalability, combined with their versatility, positions LLMs as a practical and efficient choice for researchers working across a wide range of challenges in scientometrics.

The promising results observed in this study suggest that the performance of LLMs in citation intent classification can be further improved with techniques such as chain-of-thought prompting and reasoning-focused models. These methods could enhance the models' ability to better distinguish subtle differences in intents, refining their predictions and improving overall classification accuracy.

% {\color{red} \textbf{Paris:} Discuss once again WHY we did this (the investigation) and that the LLM reaches a near-identical level of performance in scicite compared to reported literature. Mention ease of use and deployment due to the ubiquity of local LLM options as a large advantage. Think about other advantages. Mention that there is a lot of room for improvement.}

\section{Related Work}\label{sec:related_work}

% Outline:

% - First, mention the theoretical beginnings~\citep{Gar1965, MM1975, Spi1977}.

% - Then, go to first attempts to classify citations by their function in a fully automatic manner~\citep{GM2000, TST2006a}

% - Then, mention~\citep{JKH2018} that introduced the ACL-ARC dataset and \citep{PK2020} that uses a similar classification scheme with an additional layer for the COMPARE\_CONTRAST category (similarities, differences, disagreement).

% - Then go to DL methods and transformers-based models like BERT and XLNet and ensembles of them, \citep{CAZ2019, BLC2019, MRR2021, HHD2022, PVD2024}.

% - Finally, go to LLM-related architectures \citep{LSM2023, SKK2024}.

% - Close with the prompting strategies paper \citep{KPK2023} - mention the closed OpenAI models - contrast our work with this.
% \href{}{}
The theoretical beginnings of citation analysis can be traced back to foundational works such as \citeposs{Gar1965} identification of reasons for citation and \citeposs{MM1975} studies on citation function. Early annotation schemes, such as those by~\citet{Spi1977}, were later adapted by \citet{TST2006a} for supervised machine learning approaches to citation classification.

\citet{JKH2018} introduced the ACL-ARC dataset, which contains nearly $2,000$ citations from papers in the NLP field, annotated for their function with a classification scheme of six classes. \citet{PK2020} extended this classification scheme by refining the comparison class to capture similarities, differences, and disagreement. Around the same time, \citet{CAZ2019} proposed a multitask model incorporating structural information from scientific papers. They also introduced the SciCite dataset, which is significantly larger and spans multiple scientific domains, with three intent classes. 

\begin{table}[t]
    \centering
    \scriptsize
    \renewcommand{\arraystretch}{1.1}
    \begin{tabular}{lcc}
        \toprule
        \textbf{Method}                                     & \textbf{SciCite}  & \textbf{ACL-ARC}  \\
        \midrule
        Feature-rich RF~\citep{JKH2018}                     & --                & 53.00 \\
        Structural Scaffolds~\citep{CAZ2019}                & 84.00             & 67.90              \\
        SciBERT~\citep{BLC2019}                             & 85.22             & --                \\
        ImpactCite~\citep{MRR2021}                          & 88.93             & --                \\
        % VarMAE~\cite{HHD2022}                               & 86.32             & 76.50             \\
        CitePrompt~\citep{LSM2023}                          & 86.33             & 68.39             \\
        EnsIntWS~\citep{PVD2024}                            & 89.46             & --                \\
        EnsIntWoS~\citep{PVD2024}                           & 88.48             & --                \\
        MTL Finetuning (Search)~\citep{SKK2024}             & 85.25             & 64.56             \\
        MTL Finetuning (TRL)~\citep{SKK2024}                & 85.35             & 75.57             \\
        \bottomrule
        \end{tabular}
        \caption{Reported F1-Scores of notable works from the literature, sorted chronologically.}\label{tab:literature-f1}
\end{table}

\citet{BLC2019} introduced SciBERT, a BERT-based~\citep{delvin-bert} encoder language model pre-trained specifically on scientific text, which has since become the backbone of many citation intent classification methods. SciBERT has been widely adopted due to its ability to generalize across scientific domains. For example, \citet{MRR2021} introduced ImpactCite, an XLNet-based method for citation impact analysis, which was later used by \citet{PVD2024} to achieve state-of-the-art results on the SciCite dataset. Paolini et al. demonstrated the effectiveness of ensemble classifiers combining fine-tuned SciBERT and XLNet models. 
% On the ACL-ARC dataset, \cite{HHD2022} achieved state-of-the-art results by studying domain adaptation of language models using a variational masked autoencoder.

Recent research has continued to explore PLM-based methods for citation intent classification. \citet{LSM2023} used a prompt-based learning approach on SciBERT to identify citation intent, while \citet{SKK2024} achieved state-of-the-art performance on the ACL-ARC dataset by proposing a multi-task learning framework that jointly fine-tunes SciBERT on a dataset of primary interest together with multiple auxiliary datasets to take advantage of additional supervision signals.
%propose a multi-task learning framework that jointly fine-tunes SciBERT on a dataset of primary interest together with multiple auxiliary datasets to take advantage of additional supervision signals. 
\citet{KPK2023} explored various prompting and tuning strategies on SciBERT, including fixed and dynamic context prompts, and found that parameter updating with prompts improved performance. They also briefly experiment on LLMs by evaluating the zero-shot performance of GPT-3.5, which performed well on their recently introduced ACT2 dataset but poorly on the ACL-ARC dataset. However, GPT-3.5 was not evaluated on SciCite. In contrast, our work is the first to focus entirely on evaluating and fine-tuning numerous open-weight, general-purpose LLMs without using models pre-trained explicitly and exclusively for the scientific domain. 

Table~\ref{tab:literature-f1} provides a summary of the reported F1-scores of notable works in the literature, highlighting the progression of methods and datasets. SciBERT-based methods dominate the field, while our work is the first to examine LLMs on this task without any reliance to SciBERT.

\section{Conclusions}

This study investigated open Large Language Models (LLMs) for citation intent classification, demonstrating their viability with in-context learning, particularly when guided by optimized prompting strategies. We found that supervised fine-tuning with minimal data significantly boosts performance; notably, our fine-tuned Qwen 2.5 14B model achieved relative F1-score improvements of approximately $8$\% on SciCite and $4.3$\% on ACL-ARC over strong instruction-tuned baselines, reaching performance levels competitive with specialized systems. The detailed insights from our prompting parameter experiments, combined with our openly available evaluation framework and models, aim to facilitate further research and application of LLMs in scientometrics.

% This study explored the capabilities of open LLMs for citation intent classification, focusing on their performance with and without task-specific training. We found that while instruction-tuned LLMs perform reasonably well through in-context learning, supervised fine-tuning with minimal data leads to significant improvements. Additionally, our experiments on prompting-related parameters offer practical insights into optimizing model performance. To support further research, we provide our evaluation platform and models to the research community. 
% \input{sections/8_limitations}

\section*{Acknowledgments}

This work was co-funded by the EU Horizon Europe projects SciLake (GA: 101058573) and GraspOS (GA: 101095129). Part of this work utilized Amazon's cloud computing services, which were made available via GRNET under the OCRE Cloud framework, providing Amazon Web Services for the Greek Academic and Research Community. The authors wish to also thank Dimitris Roussis and Leon Voukoutis, members of Athena RC's Institute for Speech and Language Processing for their assistance on key points of this work.

% \newpage

\bibliography{custom}

\newpage

\appendix
% \counterwithin{figure}{section}
% \counterwithin{table}{section}
% \renewcommand\thefigure{\thesection\arabic{figure}}
% \renewcommand\thetable{\thesection\arabic{table}}

% \input{appendices/a_system_prompts}
% \input{appendices/b_statistical_test}
% \input{appendices/c_models}
% \input{appendices/d_top_configurations}

\end{document}